\documentclass[conference]{IEEEtran}
\IEEEoverridecommandlockouts
\usepackage{cite}
\usepackage{amsmath,amssymb,amsfonts}
\usepackage{algorithmic}
\usepackage{graphicx}
\usepackage[whole]{bxcjkjatype}
\usepackage{url}
\usepackage{textcomp}
\usepackage{xcolor}
\def\BibTeX{{\rm B\kern-.05em{\sc i\kern-.025em b}\kern-.08em
    T\kern-.1667em\lower.7ex\hbox{E}\kern-.125emX}}

\usepackage{tcolorbox}
\tcbuselibrary{raster,skins,breakable}
\tcbuselibrary{xparse} 
\DeclareTColorBox{brekableitembox}{ o m O{.5} O{} }%
  {empty, left=2mm, right=2mm, top=-2mm, bottom=0.2mm, attach boxed title to top left={xshift=1.2em},
  boxed title style={empty,left=-2mm,right=-2mm}, colframe=black, coltitle=black, coltext=black, breakable,  
  underlay unbroken={\draw[black,line width=#3pt]
    (title.east) -- (title.east-|frame.east) -- (frame.south east) -- (frame.south west) -- (title.west-|frame.west) -- (title.west); },
  underlay first={\draw[black,line width=#3pt](title.east) -- (title.east-|frame.east) -- (frame.south east) ;
    \draw[black,line width=#3pt] (frame.south west) -- (title.west-|frame.west) -- (title.west); },
  underlay middle={\draw[black,line width=#3pt](frame.north east) -- (frame.south east) ;
    \draw[black,line width=#3pt](frame.south west) -- (frame.north west) ;},
  underlay last={\draw[black,line width=#3pt](frame.north east) -- (frame.south east) -- (frame.south west) -- (frame.north west) ;}, IfValueTF={#1}{title=~~#2~~〈#1〉}{title=~~#2~~},#4}

\begin{document}
    
\title{Indexing Economic Fluctuation Narratives \\from Keiki Watchers Survey
}

\makeatletter
\newcommand{\linebreakand}{%
  \end{@IEEEauthorhalign}
  \hfill\mbox{}\par
  \mbox{}\hfill\begin{@IEEEauthorhalign}
}
\makeatother

\author{\IEEEauthorblockN{Eriko Shigetsugu}
\IEEEauthorblockA{\textit{School of Engineering }}
\IEEEauthorblockA{\textit{Hokkaido University} \\
shigetsugu.eriko.j0@elms.hokudai.ac.jp}
\and
\IEEEauthorblockN{Hiroki Sakaji}
\IEEEauthorblockA{\textit{Faculty of Information Science and Technology}}
\IEEEauthorblockA{\textit{Hokkaido University} \\
sakaji@ist.hokudai.ac.jp}
\linebreakand
\IEEEauthorblockN{Itsuki Noda}
\IEEEauthorblockA{\textit{Faculty of Information Science and Technology}}
\IEEEauthorblockA{\textit{Hokkaido University} \\
i.noda@ist.hokudai.ac.jp}
}

\maketitle

\begin{abstract}
In this paper, we design indices of economic fluctuation narratives derived from economic surveys.
Companies, governments, and investors rely on key metrics like GDP and industrial production indices to predict economic trends.
However, they have yet to effectively leverage the wealth of information contained in economic text, such as causal relationships, in their economic forecasting.
Therefore, we design indices of economic fluctuation from economic surveys by using our previously proposed narrative framework.
%
From the evaluation results, it is observed that the proposed indices had a stronger correlation with cumulative lagging diffusion index than other types of diffusion indices.
%

\end{abstract}


\section{Introduction}
Companies, governments, and investors need to predict economic trends to forecast revenue, product sales, commodity prices, and stock prices.
The Diffusion Index\footnote{The diffusion index is a summary measure designed to facilitate the analysis and forecast of business cycles by combining the behavior of a group of economic indicators that represent widely differing activities of the economy, such as production
and employment, and that correspond closely to turning points.
\url{http://www.esri.cao.go.jp/en/stat/di/di2e.htm}} is one of the indices related to economic trends; it is computed every month and provides information on economic trends during the prior period.
The diffusion index is also used in corporate strategy formulation and is highly valuable.
In recent years, research on economic analysis using non-traditional data, such as satellite imagery, text data, and location information, has been increasing. Some studies use satellite imagery data to predict wheat harvests, while others employ text data to target uncertainties that were previously unaddressable. These studies have just begun and have not yet become entrenched in economic analysis.
Among non-traditional data, text, in particular, has been attracting attention. However, many studies that focus on large amounts of text-primarily involve counting the number of words included in a dictionary or using topic analysis methods such as LDA (Latent Dirichlet Allocation). Therefore, the novelty of analytical methods is not highly emphasized; rather, greater importance is placed on the subjects of analysis and the formulation of problem settings.
The text contains various information besides words and topics, and among these, we are focusing on causal relationships represented in sentence-structures. For example, in the sentence ``Due to the impact of rising prices, the number of overseas travelers is decreasing,'' the cause is ``rising prices,'' and the effect is ``the number of overseas travelers is decreasing.'' The causal relationships that appear in the text are those recognized by the person who wrote the text. Therefore, by collecting these causal relationships from the text and processing them appropriately, we believe it is possible to grasp the economy as perceived by the writer through these causal relationships.
If we can extract causal relationships from text and employ an appropriate framework capable of economic analysis, it would provide valuable insights for new economic studies. 

Previously, we proposed a framework focused on climate change, where we extracted causal relationships from newspaper articles to design a narrative index. In this paper, we design indices of economic fluctuation narratives derived from economic surveys.
The contributions of this study are threefold. First, we demonstrate the versatility of a previously developed framework by adapting and applying it to a new dataset, the Keiki Watchers Survey, showcasing its generalizability across different types of data. Second, we provide a novel perspective by extracting causal information from the judgment reasons and additional explanations in the Keiki Watchers Survey and analyzing their relationship with the Diffusion Index (DI). While existing research has focused on predicting current economic assessments in the survey, no prior studies have utilized causal information to explore how explanatory texts relate to the DI, making this analysis a fresh contribution. Third, we evaluate whether economic patterns, such as business cycles, can be effectively captured and interpreted using data from the Keiki Watchers Survey. 

To this end, we constructed a narrative index to systematically analyze and interpret the extracted causal information in relation to economic indicators.

\section{Related Work}
As related works of causal extraction, Inui et al. proposed a method for acquiring causal relations ({\it cause}, {\it effect}, {\it precond} and {\it means}) from  a complex sentence containing a Japanese resultative connective
 ため({\it tame}) \cite{inui2005acquiring}.
Khoo et al. proposed a method for extracting cause-effect information from a newspaper text and a method for extracting causal knowledge from a medical database by applying patterns made by hand\cite{khoo}\cite{khoo2}.
In these researches, {\it cause} and {\it effect} etc. need to be contained together in the
same sentence.
%
These studies primarily focus on extracting causal relations within single sentences. Based on this foundation, our previous research constructed a framework that extracts causal relations from individual sentences and connects these relations across multiple sentences or topics to provide a more comprehensive understanding of causality. In the current study, we apply this framework to a different dataset, demonstrating its adaptability and effectiveness in analyzing economic textual data.
As relevant studies on sentiment analysis for financial texts, Koppel et al. proposed a method for classifying news stories about a company according to its apparent impact on the performance of the company's stock\cite{koppel2006good}.
In general, articles concerning business performance imply content that influences the stock price.
However, even if a company's business performance is good, its stock price will not rise if its main cause is unrelated to its core business (e.g., profit from stock sales).
The expressions are extracted manually in their research, and the method needs dictionaries.

In recent years, researchers in economics and finance have increasingly used BERT to analyze large volumes of textual data, such as corporate financial statements and patent information, for tasks like text classification, information extraction, and financial data substantiation. In climate change-related analyses, Bingler et al. \cite{bingler2022cheap} extracted information on companies' progress in addressing climate change and assessed their scores, while Kölbel et al. \cite{kolbel2020ask} developed a sentiment index for physical and transition risks and conducted empirical research on its relationship with individual companies' CDS spreads. Thus, previous studies have leveraged extensive datasets to build classification models on climate change-related financial and economic topics, created news indicators and score ratings from texts, and analyzed stock returns, corporate behavior, and economic activity.

\section{Keiki Watchers Survey}

One of the novelties of this research lies in applying the method proposed by Sakaji and Kaneda \cite{sakaji2023indexing} to a different dataset and conducting experiments based on it. Specifically, the dataset used is the ``Keiki Watchers Survey\footnote{\url{https://www5.cao.go.jp/keizai3/watcher/watcher_menu.html}}.'' This chapter provides a detailed explanation of this survey data.

The keiki Watchers Survey is a monthly questionnaire survey published by the Cabinet Office of Japan, aimed at grasping the current state and trends of the Japanese economy from various perspectives such as local residents, households, businesses, and employment. This survey is conducted in Japanese, where respondents evaluate the current month's economic conditions compared to the previous month on a five-point scale. Additionally, respondents provide their industry, the reasoning behind their assessment, and further explanations supporting that reasoning. The survey data is a valuable information source reflecting the respondents' actual voices, allowing for a detailed analysis of economic trends. In this study, we analyze survey data spanning from January 2011 to December 2023. We specifically extract three elements—judgment reasons, additional explanations, and response dates—as the foundation for our analysis. The additional explanations, in particular, contain detailed textual information related to the judgment reasons, positioning them as suitable data for in-depth analysis of the economic situation and for extracting causal relationships. Table \ref{tab:Keiki_watchers_survey} provides examples of typical responses from the Keiki Watchers Survey, demonstrating the structure and types of data collected, including the industry, judgment reason, and additional explanations. These examples help illustrate the richness and depth of the survey data, which serve as the basis for our subsequent analysis.

\begin{table*}[htbp]
\centering
\caption{Sample Responses from the Keiki Watchers Survey}
\begin{tabular}{|c|l|l|p{6.5cm}|}
\hline
\textbf{Economic Condition} & \textbf{Occupation} & \textbf{Judgment Reasons} & \textbf{Additional Explanations} \\ \hline

○ & Other Specialty Store (Liquor) (Owner) & Number of Visitors & As the weather has started to warm slightly at the end of February, customer traffic has improved. \\ \hline
◎ & Convenience Store (Area Manager) & Unit Price Movement & The decline in customer numbers has stopped, and the year-on-year performance of existing stores continues to improve, as seen last month. \\ \hline
× & Electronics Retail Store (Manager) & Sales Volume Movement & Due to early demand for eco-point electronic goods, the sales volume has now slowed. \\ \hline
\end{tabular}
\label{tab:Keiki_watchers_survey}
\end{table*}

\begin{table}[ht]
    \centering
    \caption{List of 13 Topics from the Keiki Watchers Survey}

    \begin{tabular}{p{8cm}}
        \hline
        Competitor Behavior, Customer Behavior, Employment Form Movement, Job Offer Movement, Job Seeker Movement, Neighboring Company Behavior, Number of Hires, Number of Visitors, Order \& Sales Price Movement, Order \& Sales Volume Movement, Sales Volume Movement, Trading Partner Behavior, Unit Price Movement. \\
        \hline
    \end{tabular}
    \label{tab:topics}
\end{table}

\section{Methodology}
Sakaji and Kaneda \cite{sakaji2023indexing} proposed a framework for analyzing newspaper articles. In this framework, topics were predicted from the articles using BERT, while causal information was independently extracted from the same articles. The extracted causal relations were then used to construct narratives across topics, which were subsequently indexed to enable deeper analysis of the articles. In contrast, this study does not employ topic prediction. Instead, it adopts the ``judgment reasons'' provided in the Keiki Watchers Survey as pre-defined topics. This approach is reasonable because the judgment reasons are already categorized into 13 types (see Table \ref{tab:topics}), eliminating ambiguity in topic setting. As a result, this method is expected to improve the accuracy of causal relationship extraction without the need for topic prediction.

In this chapter, we provide a detailed explanation of the process of extracting causal information, constructing narratives, and indexing them using the new dataset, the Keiki Watchers Survey. The overall process is illustrated in Figure \ref{fig:flow}. 

\begin{figure}
    \centering
    \includegraphics[width=1\linewidth]{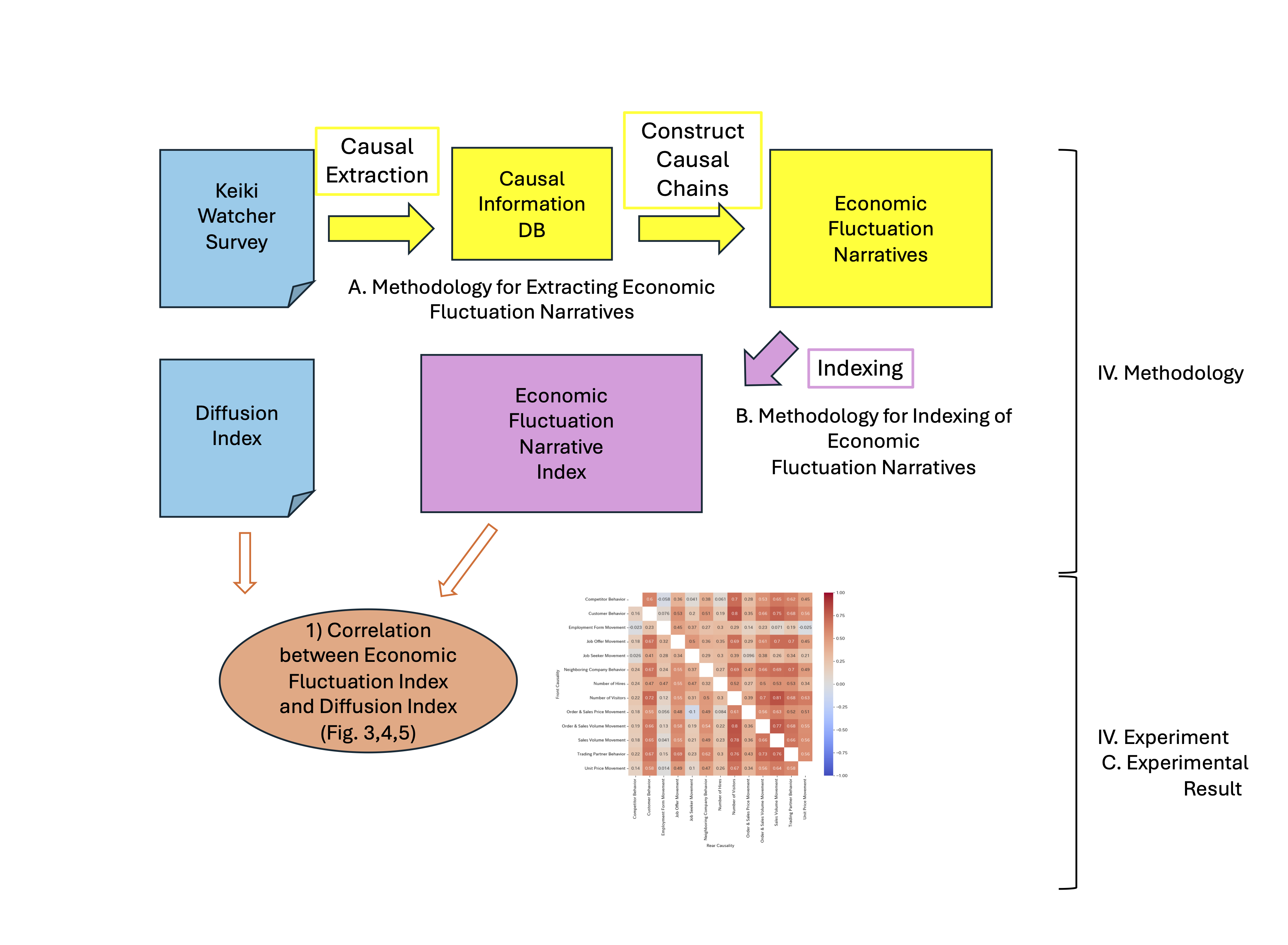}
    \caption{Overview: The methodology begins with extracting causal information from the Keiki Watchers Survey, followed by constructing causal chains to form economic fluctuation narratives. These narratives are indexed to calculate the Economic Fluctuation Narrative Index, which is then analyzed for correlations with the Diffusion Index (DI). The results are visualized using heatmaps to highlight the relationships between the narratives and the DI.}
    \label{fig:flow}
\end{figure}
\subsection{Methodology for Extracting Economic Fluctuation Narratives}
Building on the method proposed by Sakaji and Kaneda \cite{sakaji2023indexing}, we extracted Economic Fluctuation Narratives using the following steps. In Step 1, we performed causal extraction (left yellow arrow), followed by the extraction of economic fluctuation narratives in Step 2 (right yellow arrow).

\subsubsection{Step 1} 

Causal extraction was applied to the ``additional explanations'' section of the dataset while preserving the assigned topics. First, we determined whether the additional explanations contained causal relationships using the algorithm proposed by Sakaji and Masuyama \cite{sakaji2011news}. Next, we extracted pairs of expressions representing the logical relationships between cause and effect, using the causal extraction method proposed by Sakaji {\it et al.} \cite{sakaji2008extracting}, and recorded them in a causal information database. Specifically, causal terms and resulting terms were extracted from sentences using keywords that indicate causal relationships, such as ``due to,'' ``because of,'' ``as a result,'' ``therefore,'' and similar phrases. This entire process was executed using programs developed in the referenced studies, with no manual intervention.
\subsubsection{Step 2}Economic Fluctuation Narratives were extracted through Causal Chain Construction.
By linking the causal information obtained in Step 1 with their respective topics, we organized sets of causal information that span across different topics. In this process, we combined causal information stored in the database between any two topics, allowing us to extract Economic Fluctuation Narratives—a series of interconnected events that describe how economic factors influence one another over time. Specifically, these narratives assume a temporal relationship where past economic fluctuations affect current ones, allowing us to integrate past causal events with present effect events. This process aligns with the causal chain construction method proposed by Sakaji and Kaneda \cite{sakaji2023indexing}, as well as by Izumi {\it et al.} \cite{izumi2020economic}. Figure ~\ref{fig:chain_example} illustrates the linking process, where the Job Offer Movement is treated as the cause event and the Sales Volume Movement as the effect event. First, we calculated the cosine similarity between the causal expression corresponding to the current Sales Volume Movement (effect event) and the effect expression corresponding to the past Job Offer Movement (cause event) (blue line). If the cosine similarity for a pair exceeds the threshold of 0.5, they are linked. Finally, if a pair is linked, it is considered that a narrative exists between the topics to which the cause and effect events belong. In the example, the cosine similarity was 0.538, which was above the threshold, and since the cause event occurred before the effect event, it was determined that a narrative exists between the two expressions. The cosine similarity was calculated by vectorizing the causal and effect expressions using a Japanese BERT model.

\begin{figure}
    \centering
    \includegraphics[width=1\linewidth]{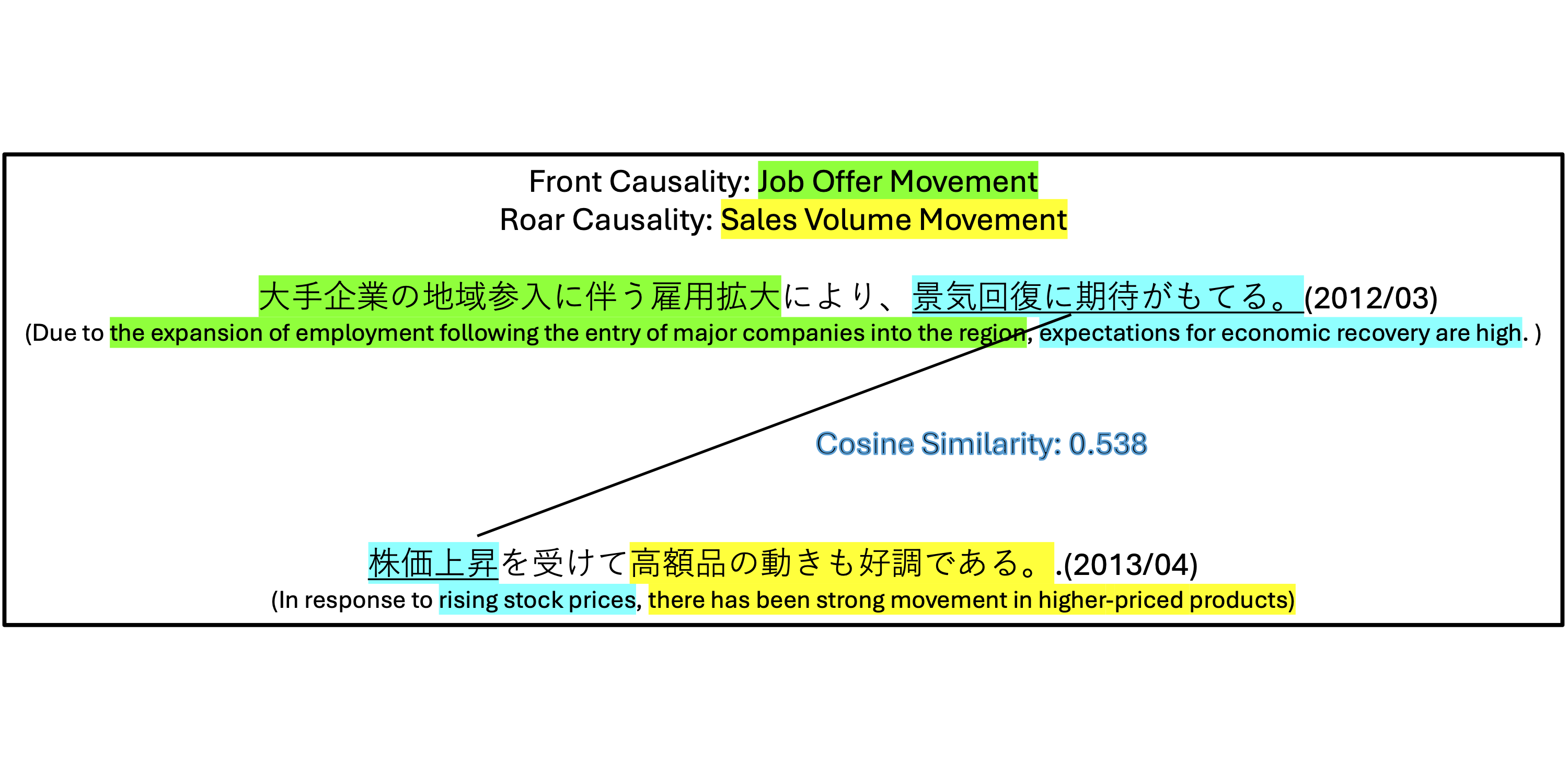}
    \caption{Chain Example: This figure illustrates an example of constructing a causal chain. When two explanation texts from different topics contain causal relationships and the cosine similarity between the result expression of the earlier sentence and the cause expression of the later sentence exceeds 0.5, they are considered part of the same chain. In this example, the earlier topic, "Job Offer Movement," and the later topic, "Sales Volume Movement," are linked, forming a narrative.}
    \label{fig:chain_example}
\end{figure}

As described by Sakaji and Kaneda \cite{sakaji2023indexing}, this corresponds to the assumption that:\\
\begin{itemize}
    \item A causal relationship ``If $X$, then $Y$'' exists in a survey on topic $A$.
    \item A causal relationship ``If $Y'$, then $Z$'' exists in a later survey on topic $B$, conducted after the survey on topic $A$.
    \item There is a high degree of similarity between $Y$ and $Y'$.
\end{itemize}

According to the transitivity rule, if these three conditions are satisfied, then:
\begin{enumerate}
    \item ``If $X$, then $Z$'' holds, and
    \item A narrative exists connecting topic $A$, where the causal event $X$ belongs, to topic $B$, where the effect event $Z$ belongs.
\end{enumerate}

\subsection{Methodology for Indexing of Economic Fluctuation Narratives}

Next, we explain the process of creating indices for economic fluctuation narratives that takes into account the causal relationships between topics (the purple section in Figure \ref{fig:flow}). This method follows the approach proposed by Sakaji and Kaneda \cite{sakaji2023indexing}.

In the first phase, we indexed the ``Economic Fluctuation Narratives between two topics at different time points,'' as shown in Equation \eqref{indexing}, and calculated a monthly index. To avoid bias where more recent news tends to have stronger connections to past events, we gradually reduced the weight of older causal relationships over time. Specifically, we used a logistic function to model the decay, with the parameters $a$ and $b$ set based on the decay period of the news. In this study, we set $a = 0.02$ and $b = 0.065$, such that the weights halve after five years.
\begin{equation}
    \mathit{Monthly\_Index(m)} = \sum_{j=0}^{M(m)}\sum_{i=0}^{L(j)} \frac{1}{1+ae^{bd}}\cos(\Vec{i}_{t-d}\cdot\Vec{j}_t)\label{indexing}
\end{equation}
\begin{equation}
    \cos(\Vec{i}_{t-d}\cdot\Vec{j}_t) = \frac{\Vec{i}_{t-d}\cdot\Vec{j}_t}{|\Vec{i}_{t-d}||\Vec{j}_t|}
\end{equation}
Here,
\begin{align*}
M(m)  : & \text{ set of causal chains included in month } m. \\
L(j) : & \text{ set of cause event } \vec{i}_{t-d} \text{ connected to result event }\\ 
    & \vec{j}_t. \\
t - d : & \text{ observation point of the cause event leading to }\\ &\text{the result event.}\text{(}d > 0\text{)} \\
t : & \text{ observation point of the result event included in } \\ & \text{month } m. \\
d : & \text{ time difference (in days) between the cause event}\\ & \text{ and the result event.}
\end{align*}

In the second phase, we considered the causal relationships between any two different topics and specified the order of cause and effect events. This resulted in the aggregation of 156 time-series datasets, calculated as $13 \text{ topics} \times 12 \text{ topics} = 156$. We refer to this as the Economic Fluctuation Narrative Indices.

\section{Experiment}
This section presents the results of evaluating the Economic Fluctuation Narrative Indices. We first briefly introduce the models and keywords used for calculating the indices, followed by detailed results of their correlations with existing economic indicators.
\subsection{Models}
To vectorize the Japanese expressions for cause events and effect events in the creation of Economic Fluctuation Narratives, we used the Japanese sentence-BERT model v.2\footnote{\url{https://huggingface.co/sonoisa/sentence-bert-base-ja-mean-tokens-v2}}, which is based on the Sentence BERT model \cite{reimers2019sentence}. This model is pre-trained specifically for Japanese text and excels at context-based text vectorization, producing 768-dimensional vectors to represent Japanese economic narratives with high accuracy.
\subsection{Clue Expressions}
In this experiment, we used keywords to extract economic fluctuation narratives. Examples include ``Because of,'' ``Due to,'' and ``affected by.'' We employed a total of 41 Japanese keywords to analyze the Keiki Watchers Survey. The English translations of these keywords are listed in Table \ref{tab:clue_expressions}. Duplicates, such as instances where different Japanese expressions translate into the same English word or where commas are present or absent, have been omitted. The original Japanese version of the keywords is provided in Appendix Table \ref{tab:clue_expressions_jp}.

\begin{table}[ht]
    \centering
    \caption{Clue Expressions}

    \begin{tabular}{p{8cm}}
        \hline
        Against the backdrop of, Due to, Because of, Accompanying, Accompanied by, Reflecting, Triggered by, Supported by, By, Affected by, In response to, Since, Influenced by, Being affected by, For this reason, Therefore, As a result. \\
        \hline
    \end{tabular}

    \label{tab:clue_expressions}
\end{table}

\subsection{Experimental Results}



To evaluate how well the Economic Fluctuation Narrative Indices tracks actual economic trends, we examined the correlation between the time-series data of each narrative's index and the time-series data of the diffusion index (DI). The DI, which measures economic conditions, is divided into three categories: Leading DI, Coincident DI, and Lagging DI, each serving to reflect different phases of the economy. The DI is calculated by comparing the values of selected series for a given month with their values three months earlier. An increase is indicated as ``+'', no change as ``0'', and a decrease as ``-''. For each category—Leading, Coincident, and Lagging DI—the percentage of expanding series (indicated by ``+'') relative to the total number of selected series is computed as the DI value, with series showing no change (indicated by "0") counted as 0.5. This method helps mitigate the effects of irregular fluctuations and captures trends similar to those shown by a three-month moving average. The number of selected series is 11 for the Leading DI, 10 for the Coincident DI, and 9 for the Lagging DI, totaling 30 series. Additionally, there are six types of DI in total, including the cumulative values for each category: Cumulative Leading DI, Cumulative Coincident DI, and Cumulative Lagging DI. Cumulative DIs are derived by subtracting 50 from the monthly DI values and summing these monthly values over time. These cumulative indices are used to visualize the cyclical trends of the DI. 

The Leading DI includes indicators aimed at forecasting future economic trends, such as the ``Inventory Ratio of Final Demand Goods (Inverse Cycle),'' ``New Job Offers (Excluding Fresh Graduates),'' ``Floor Space of New Housing Starts,'' ``Tokyo Stock Price Index (TOPIX),'' and ``Consumer Confidence Index (Seasonally Adjusted).'' The Coincident DI comprises indicators that reflect the current state of economic activity, including the ``Shipments of Industrial Production Goods Index,'' ``Index of Overtime Work Hours (Surveyed Industries Total),'' ``Commercial Sales Value (Retail and Wholesale),'' ``Operating Profits (All Industries),'' and ``Ratio of Active Job Openings to Applicants (Excluding Fresh Graduates).'' The Lagging DI consists of indicators that respond to economic activity with a delay, such as the ``Index of Regular Employment (Surveyed Industries Total),'' ``Household Consumption Expenditure (Working Households),'' ``Corporate Tax Revenue,'' ``Unemployment Rate (Inverse Cycle),'' and ``Consumer Price Index (Excluding Fresh Foods).''

To determine which narratives show the strongest correlation with the various types of DI, we calculated the Pearson correlation coefficients between the time-series data for each narrative index and the DI, and visualized the results in a heatmap (orange arrows in Figure \ref{fig:flow}). Overall, the highest correlation was observed with the cumulative lagging DI (Figure \ref{fig:cumu_lagging}). For comparison, we also present heatmaps for the lagging DI in Figure \ref{fig:lagging} and the cumulative coincident DI in Figure \ref{fig:cumu_coin}. In these heatmaps, the rows represent front expressions, and the columns represent rear expressions. For example, in the heatmap for the cumulative lagging DI (Figure \ref{fig:cumu_lagging}), the Pearson correlation between the narrative index, where the front event is the ``Number of Visitors'' and the rear event is the ``Sales Volume Movement,'' is 0.81.

Additionally, Figure \ref{fig:cumu_lagging_time_series}, Figure \ref{fig:lagging_time_series}, Figure \ref{fig:lagging_coin_time_series} each show the normalized time-series data of the top four narratives with the highest correlation coefficients from Figure \ref{fig:cumu_lagging}, Figure \ref{fig:lagging}, Figure \ref{fig:cumu_coin}, respectively, along with the normalized time-series data of the corresponding DI.

\begin{figure*}
    \centering
    \includegraphics[width=0.65\linewidth]{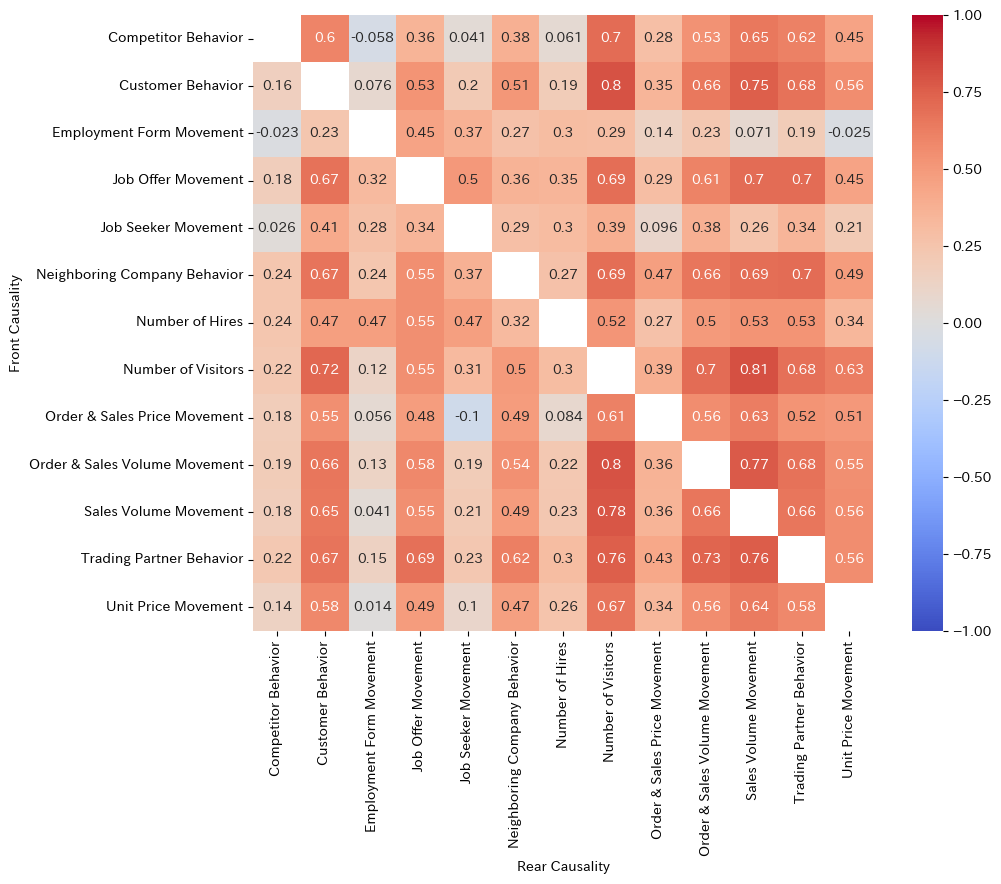}
    \caption{Heatmap of Pearson Correlations between Economic Narratives and Cumulative Lagging Diffusion Index: The heatmap illustrates Pearson correlations between indices derived from economic narratives and the Cumulative Lagging Diffusion Index (DI). Rows represent front causality (cause expressions), and columns represent rear causality (result expressions). For example, the index calculated from the narrative linking "Number of Visitors" as the front causality and "Sales Volume Movement" as the rear causality shows a Pearson correlation of 0.81 with the Cumulative Lagging DI. This highlights the strong relationship between specific economic causal chains and the DI.}
    \label{fig:cumu_lagging}
\end{figure*}
\begin{figure*}
    \centering
    \includegraphics[width=0.65\linewidth]{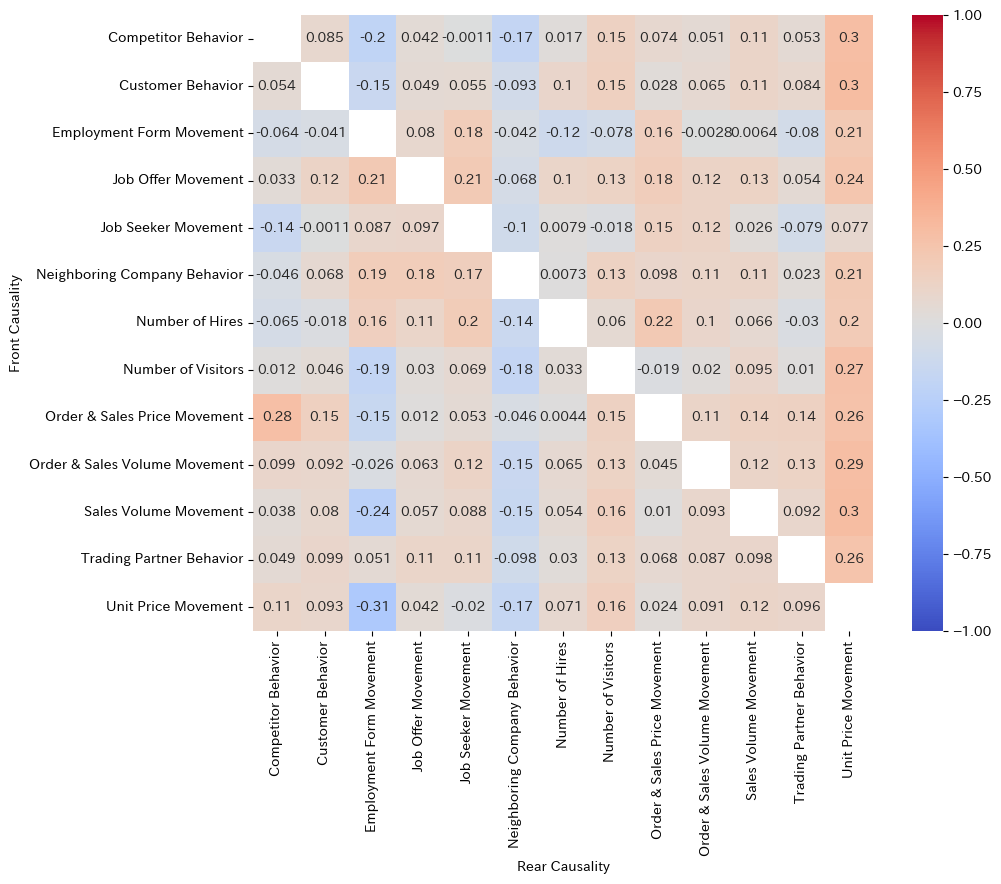}
    \caption{Heatmap of Pearson Correlations between Economic Narratives and Lagging Diffusion Index}
    \label{fig:lagging}
\end{figure*}
\begin{figure*}
    \centering
    \includegraphics[width=0.65\linewidth]{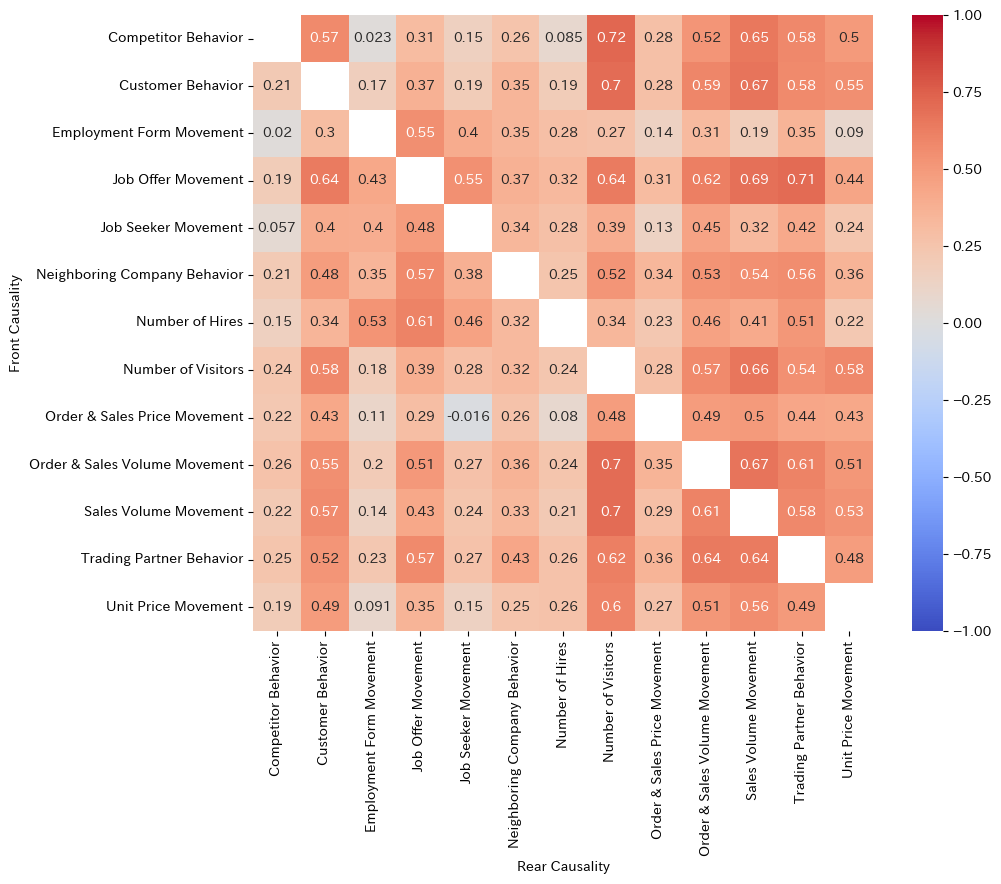}
    \caption{Heatmap of Pearson Correlations between Economic Narratives and Cumulative Coincident Diffusion Index }
    \label{fig:cumu_coin}
\end{figure*}
\begin{figure}
    \centering
    \includegraphics[width=1\linewidth]{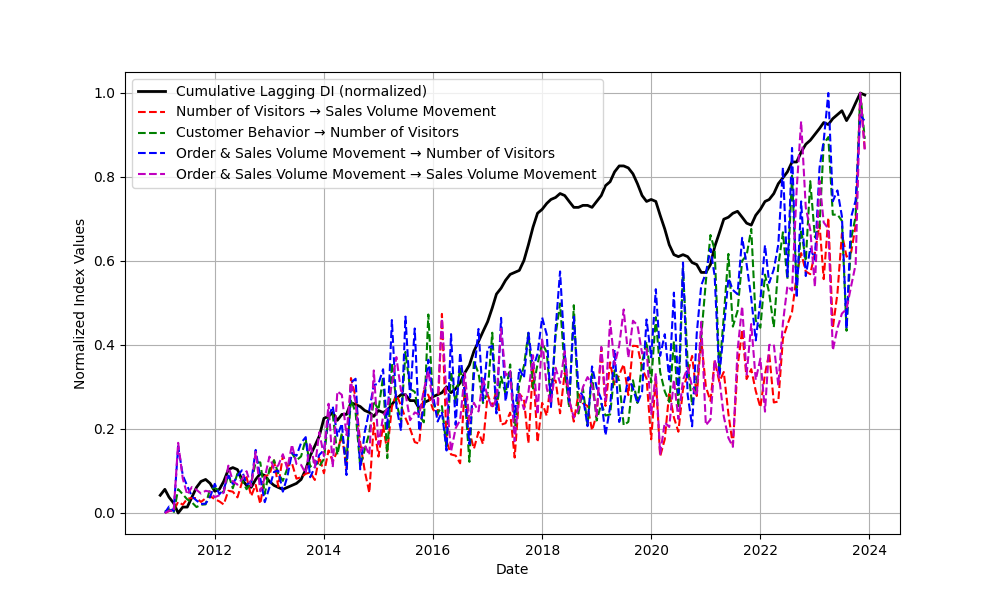}
    \caption{A time-series graph of the Cumulative Lagging DI and the top four narratives with the highest correlation coefficients, with all indicators normalized..}
    \label{fig:cumu_lagging_time_series}
\end{figure}
\begin{figure}
    \centering
    \includegraphics[width=1\linewidth]{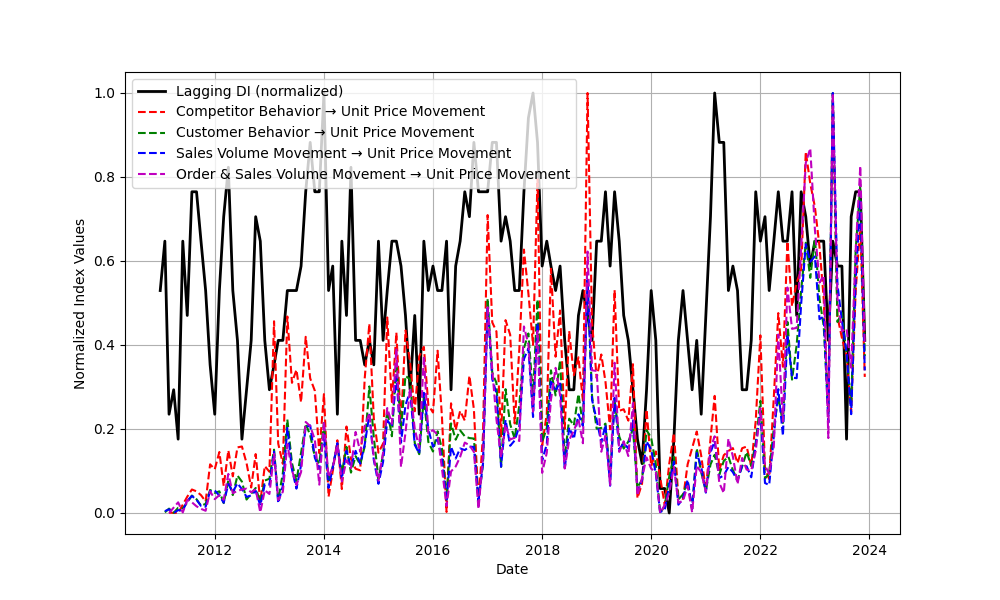}
    \caption{A time-series graph of the Lagging DI and the top four narratives with the highest correlation coefficients, with all indicators normalized.}
    \label{fig:lagging_time_series}
\end{figure}
\begin{figure}
    \centering
    \includegraphics[width=1\linewidth]{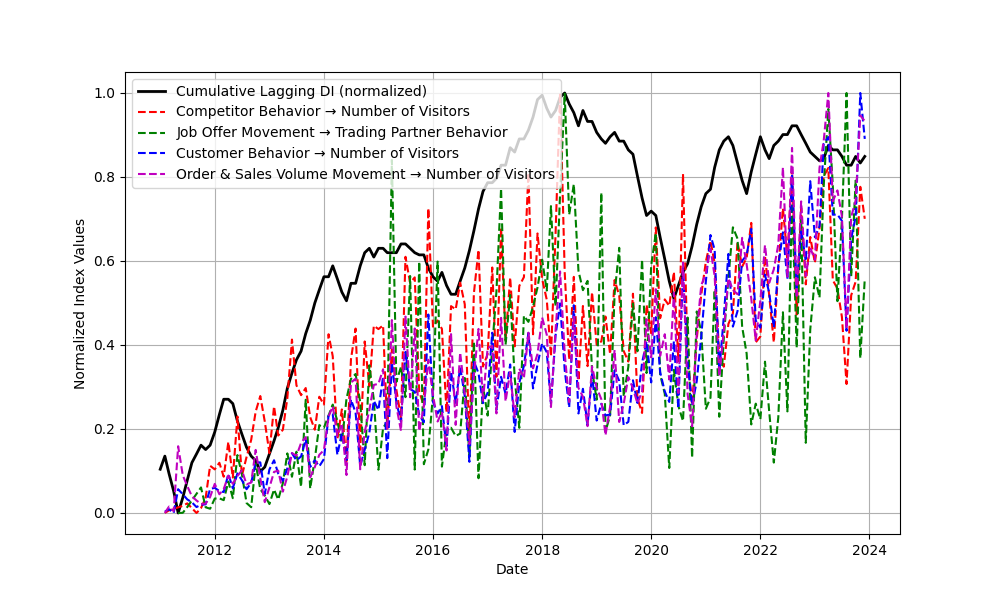}
    \caption{A time-series graph of the Coincident DI and the top four narratives with the highest correlation coefficients, with all indicators normalized.}
    \label{fig:lagging_coin_time_series}
\end{figure}

\section{Discussion}
The following points were derived from the correlation analysis between the Economic Fluctuation Narrative Indices and the DI conducted in the previous section.
\subsubsection{}
A comparison of Figure \ref{fig:cumu_lagging} and Figure \ref{fig:lagging} shows that, in this experiment, the Economic Fluctuation Narrative Indices have stronger correlations with the cumulative DI than with the regular DI. In Figure \ref{fig:lagging}, which displays the heatmap of Pearson correlation coefficients with the lagging DI, the highest value is 0.3. In contrast, Figure \ref{fig:cumu_lagging}, which shows the heatmap for the cumulative lagging DI, has a peak value of 0.81. Additionally, by comparing the normalized time series data of the indices shown in Figure \ref{fig:cumu_lagging_time_series} and Figure \ref{fig:lagging_time_series}, it is clear that the cumulative lagging DI is more closely followed. This can be explained by the fact that the Keiki Watchers Survey examines changes relative to previous periods, leading to a stronger correlation with cumulative values.
\subsubsection{}

By comparing Figure \ref{fig:cumu_lagging} and Figure \ref{fig:cumu_coin}, we can see that the Economic Fluctuation Narrative Indices in this experiment have higher correlations with the cumulative lagging DI than with the cumulative coincident DI. Similarly, comparing the normalized time series data shown in Figure \ref{fig:cumu_lagging_time_series} and Figure \ref{fig:lagging_coin_time_series} reveals that the cumulative lagging DI is more closely followed than the cumulative coincident DI. To reiterate, the coincident DI moves in step with economic trends and is typically used to assess the current state of the economy. In contrast, the lagging DI follows economic movements and is often used to identify turning points. These results suggest that the narratives in this experiment are more sensitive to detecting shifts in economic trends. Furthermore, given the nature of the Economy Watchers Survey, which tends to emphasize indicators related to household performance, small and medium-sized enterprises, and job openings, it is natural that the cumulative lagging DI exhibited the highest correlation, followed by the cumulative coincident DI. This result is noteworthy as it reflects how the survey's characteristics are mirrored in the experimental outcomes.

\subsubsection{}
Figure \ref{fig:cumu_lagging} shows that, for most causal events, narratives with customer traffic and sales volume movements as outcome events have particularly strong correlations with the DI. For example, in Figure \ref{fig:cumu_lagging}, the narrative ``customer traffic → sales volume movement'' has the highest Pearson correlation of 0.81 with the cumulative lagging DI, followed by ``customer behavior → customer traffic movement'' and ``order and sales volume movement → customer traffic movement,'' both with a value of 0.80. This trend can also be observed in Figure\ref{fig:cumu_coin}, which shows the heatmap of the Economic Fluctuation Narrative Index and the cumulative coincident DI. Examples of the causal chains from the three narratives mentioned above are listed in Table \ref{tab:examples_of_chains}. These results indicate that the experiment successfully extracted common and predictable correlations, such as the direct connection between an increase in customer traffic and changes in sales volume. Additionally, it suggests that these movements are key topics in economic fluctuations.


\begin{table*}[ht]
\centering
\caption{Examples of causal chains: Topic A refers to ``Number of Visitors,'' Topic B to ``Sales Volume Movement,'' Topic C to ``Customer Behavior,'' and Topic D to ``Order \& Sales Volume Movement.''}
\begin{tabular}{|c|p{3.5cm}|p{3.5cm}||c|p{3.5cm}|p{3.5cm}|}
\hline
\multicolumn{3}{|c||}{\textbf{Front Causality}} & \multicolumn{3}{c|}{\textbf{Rear Causality}} \\ \hline
\textbf{Topic} & \textbf{Cause} & \textbf{Effect} & \textbf{Topic} & \textbf{Cause} & \textbf{Effect} \\ \hline
A & 政権交代による景気回復の期待感 (Expectations of economic recovery due to the change of government) & 消費税増税前の駆け込み需要がいよいよ始まった (The rush demand before the consumption tax hike has finally begun.) & B & 消費税増税の話 (Talk of consumption tax hike) & 新車に乗換えを検討される客が多くなってきている. (More and more customers are considering changing to a new car.) \\ \hline

A & 台風21号と北海道胆振東部地震の影響 (The impact of Typhoon No. 21 and the Hokkaido earthquake) &国内観光客も外国人観光客も、ほとんどの予約がキャンセルとなった(Most reservations for both domestic and foreign tourists were canceled.) & B &ほぼ予約がない状況(Almost no reservations were made.) &結婚式や法人利用の宴会は、一進一退を繰り返している. (Weddings and corporate use banquets are going back and forth.) \\ \hline

C & 客に節約する様子が見られる(Customers appear to be saving money.) & 単価の高い商品やおせちなどの催事物の売上が減少している(Sales of high unit price products and special events such as Osechi are declining.) & A & 量販店での低価格品に消費が流れている影響(The impact of consumption flowing to lower-priced products at mass retailers.) & 眼鏡の購入客数も激減している. (The number of customers buying eyeglasses has also plummeted.)\\ \hline

C & 東日本大震災の建築物等の需要増加(Increased demand for buildings and other construction materials following the Great East Japan Earthquake.) & 建設業界の景気は上昇している. (The construction industry's economy is rising.) & A & ただし住宅リフォームは依然として好調(However, home remodeling remains strong.) & 前年比200％以上の伸びを示しており(More than 200\% year over year growth.)\\ \hline

D & 当社を始めとする小売チェーンの商品の多くが輸入品(Many of the products of our company and other retail chains are imported.) & 円安の影響を大きく受けている (The depreciation of the yen has had a significant impact.) & A & 円安や春節の影響(The impact of the weak yen and Chinese New Year) & アジアを含めた観光客の需要が目にみえて増加している. (Demand from tourists, including those from Asia, is increasing noticeably)\\ \hline

D & 小売業やサービス業は、売上、利益の前年割れ状態が続いている. 建設関係は一定の売上と利益を計上しているものの、前年比はやはり落ち込んでいる.(The retail and service industries continue to see year over year declines in sales and profits. Construction-related businesses are posting a certain amount of sales and profits, but are still down from the previous year) & 顧客の中には、統廃合や後継者問題で事業を継承できずに会社をたたむケースが増えてきている.販売先が減少しており、受注活動に若干の影響を及ぼしている. (An increasing number of customers are folding their companies due to consolidation and succession issues. The number of customers to whom we sell our products has been decreasing, which has had some impact on our order-taking activities.) & A & 固定客の移転や病気、銀行の店舗閉鎖 (Fixed customers are relocating, illnesses and bank branches are closing.) & 来客数が減少し、売上も減少している。 (The number of visitors is decreasing and sales are also reducing.)\\ \hline

\hline
\end{tabular}\label{tab:examples_of_chains}
\end{table*}

\section{Conclusion}
In this paper, we analyzed Keiki Watchers Survey using our narrative framework.
First, our framework extracted causality from the Keiki Watchers Survey.
Then, our framework constructed causal chains using extracted causality and Sentence-BERT.
Finally, the Economic Fluctuation Narrative Indices were calculated using constructed causal chains.
As an experiment results, we confirmed that our indices had strong correlations with cumulative coincident diffusion index.
This is because the comments from economy watchers are describing fluctuations in the economy since the last time.
In future work, we have a plan to calculate the Granger causality of each narrative indices for analyzing leading and lagging.
Additionally, we will tackle creating English narrative indices.
To achieve this, it is necessary to establish a method for extracting causal relationships from English documents.

\section*{Acknowledgment}
This work was supported by JST PRESTO Grant Number JPMJPR2267, Japan.

\bibliographystyle{IEEEtran}
\bibliography{IEEEabrv,myrefs}

\appendix
%
Table \ref{tab:clue_expressions_jp} shows the original clue expressions used in this experiment.
\begin{table}[ht]
    \centering
    \caption{Original Clue Expressions in Japanese}
    \begin{tabular}{p{8cm}}
        \hline
        を背景に, を背景に、, を受け、, ため、, に伴う, に伴い、, を反映して, をきっかけに, により、, に支えられて, によって, を反映し、, が響き、, ためで、, を受けて, から、, により, が響いた。 ,ため」, が影響した。, による。, ためで, ためだ。, を受けて、, に伴い, ため。, が響く, が響いている, が響いている。, で、, を受けております。, によります。, によっております。, ためであります。, によっています。, このため、, このため, そのため、, そのため, その結果、, この結果、\\
    \hline
    \end{tabular}
    \label{tab:clue_expressions_jp}
\end{table}
\end{document}